\documentclass{article}

\usepackage{microtype}
\usepackage{graphicx}
\usepackage{subcaption}
\usepackage{booktabs} 
\usepackage{multirow}
\usepackage{hyperref}
\usepackage{enumitem} 
\usepackage{soul} 

\usepackage[accepted]{icml2026}

\providecommand{\algorithmicrequire}{\textbf{Require:}}
\providecommand{\algorithmicensure}{\textbf{Ensure:}}
\providecommand{\algorithmiccomment}[1]{\{#1\}}
\providecommand{\algorithmicend}{\textbf{end}}
\providecommand{\algorithmicif}{\textbf{if}}
\providecommand{\algorithmicthen}{\textbf{then}}
\providecommand{\algorithmicelse}{\textbf{else}}
\providecommand{\algorithmicfor}{\textbf{for}}
\providecommand{\algorithmicforall}{\textbf{for all}}
\providecommand{\algorithmicdo}{\textbf{do}}
\providecommand{\algorithmicwhile}{\textbf{while}}
\providecommand{\algorithmicloop}{\textbf{loop}}
\providecommand{\algorithmicrepeat}{\textbf{repeat}}
\providecommand{\algorithmicuntil}{\textbf{until}}
\providecommand{\algorithmicinput}{\textbf{input}}
\providecommand{\algorithmicoutput}{\textbf{output}}
\providecommand{\algorithmicset}{\textbf{set}}

\providecommand{\algorithmicand}{\textbf{and\ }}
\providecommand{\algorithmicor}{\textbf{or\ }}
\providecommand{\algorithmicfunction}{\textbf{function}}
\providecommand{\algorithmicmain}{\textbf{main}}

\usepackage{amsmath}
\usepackage{amssymb}
\usepackage{mathtools}
\usepackage{amsthm}
\usepackage{algorithm}
\usepackage[most]{tcolorbox}

\usepackage[capitalize,noabbrev]{cleveref}

\newtcolorbox{promptbox}[1]{
  colback=blue!4,
  colframe=blue!55!black,
  coltitle=white,
  title=#1,
  fonttitle=\bfseries,
  boxrule=0.5pt,
  arc=2pt,
  left=6pt,
  right=6pt,
  top=6pt,
  bottom=6pt
}

\newtcolorbox{operatorbox}{
  colback=blue!4,
  colframe=blue!55!black,
  coltitle=white,
  title=\faUserIcon\ Operator Prompt,
  fonttitle=\bfseries\small,
  boxrule=0.5pt,
  arc=3pt,
  left=6pt, right=6pt, top=4pt, bottom=4pt,
  before skip=4pt, after skip=2pt
}

\newtcolorbox{frameresponse}{
  colback=green!4,
  colframe=green!40!black,
  coltitle=white,
  title=\faRobotIcon\ FRAMe Response,
  fonttitle=\bfseries\small,
  boxrule=0.5pt,
  arc=3pt,
  left=6pt, right=6pt, top=4pt, bottom=4pt,
  before skip=2pt, after skip=6pt
}

\newtcolorbox{coachreview}{
  colback=red!4,
  colframe=red!60!black,
  coltitle=white,
  title=\faRobotIcon\ Coach Review,
  fonttitle=\bfseries\small,
  boxrule=0.5pt,
  arc=3pt,
  left=6pt, right=6pt, top=4pt, bottom=4pt,
  before skip=2pt, after skip=6pt
}

\providecommand{\faUserIcon}{\textbf{[U]}}
\providecommand{\faRobotIcon}{\textbf{[F]}}

\newtcolorbox{inputbox}{
  colback=blue!4,
  colframe=blue!55!black,
  coltitle=white,
  title=\textbf{FRAMe Input (SYS + USER + OPERATOR)},
  fonttitle=\bfseries\small,
  boxrule=0.5pt,
  arc=3pt,
  left=6pt, right=6pt, top=4pt, bottom=4pt,
  before skip=4pt, after skip=2pt
}

\definecolor{syscolor}{HTML}{1F4E79}     
\definecolor{usercolor}{HTML}{7B3F00}    
\definecolor{operatorcolor}{HTML}{8B0000}

\theoremstyle{plain}

\theoremstyle{definition}

\theoremstyle{remark}

\usepackage[textsize=tiny]{todonotes}

\icmltitlerunning{End-to-End LLM Flight Planning with RAG-based Memory and Multi-modal Coach Agent}

\begin{document}

\twocolumn[
  \icmltitle{End-to-End LLM Flight Planning with RAG-based Memory and Multi-modal Coach Agent}



  \icmlsetsymbol{equal}{*}

  \begin{icmlauthorlist}
    \icmlauthor{Amin Tabrizian}{equal,gwu}
    \icmlauthor{Arsyi Aziz}{equal,gwu}
    \icmlauthor{Aarifah Ullah}{gwu}
    \icmlauthor{Mahyar Ghazanfari}{gwu}
    \icmlauthor{Pouria Razzaghi}{metis}
    \icmlauthor{Peng Wei}{gwu}
  \end{icmlauthorlist}

  \icmlaffiliation{gwu}{George Washington University, Washington, DC, USA 20052}
  \icmlaffiliation{metis}{Metis Solutions Technology Inc, NASA Ames Research Center, Moffett Field, CA, USA 94035}

  \icmlcorrespondingauthor{Amin Tabrizian}{amin\_tabrizian@gwu.edu}

  \icmlkeywords{Machine Learning, ICML}

  \vskip 0.3in
]



\printAffiliationsAndNotice{\icmlEqualContribution}

\begin{abstract}
Bridging the gap between human pilot intent and autonomous flight operation is critical for real-world electric vertical takeoff and landing (eVTOL) aircraft deployment. Flight planning traditionally relies on classic algorithms that struggle to incorporate flexible human preferences. We present \textbf{FRAMe}, an End-to-End Large Language Model (LLM) \textbf{F}light Planning tool with \textbf{RA}G-based Memory and \textbf{M}ulti-modal Coach Agent. Our system integrates a planner LLM with a multi-modal coach agent and retrieval augmented generation (RAG)-based memory to generate flight plans that satisfy mission constraints while aligning with human flight operator preferences. We demonstrate the system in a range of real-world-inspired scenarios of varying difficulty levels. Across four LLMs, the full FRAMe system (RAG and coach) yields the highest validity for every planner (up to 93.8\% aggregate, 99\% on Easy scenarios for the strongest planner) and shifts preference-relevant metrics in the operator-favored direction where the metric has headroom. FRAMe signifies how advanced LLMs can be deployed for human-centric mission planning, translating natural language instructions into safe, efficient, and flexible flight routes. The code is available at: \href{https://github.com/amin-tabrizian/FlightPlanningLLMs/tree/new}{github.com/amin-tabrizian/FlightPlanningLLMs}
\end{abstract}

\section{Introduction}

Advanced Air Mobility (AAM) operations are projected to grow significantly in the near future, with unmanned aircraft systems (UAS) and eVTOL vehicles performing large volumes of flights for both cargo delivery and passenger transportation. Maintaining safe and efficient flight planning within such dense and complex airspace presents a major challenge. Flight restrictions, coupled with the often subjective and context-dependent nature of mission-specific requirements, make it difficult to define rigid, mathematical objectives that fully capture operational intent.

Classical path planning algorithms, such as A$^*$ and rapidly-exploring random trees (RRT, RRT$^*$) \cite{RRT_survey, lavalle2006planning}, can compute optimal, obstacle-free trajectories; however, they depend on explicitly defined mathematical objectives and constraints, which limits their ability to capture nuanced human pilot preferences, such as trading off flight duration against waypoint complexity. As the scale and complexity of eVTOL operations continue to grow, there is an increasing need for automated planning systems capable of interpreting such nuanced mission intent.

One promising technology to bridge this gap is to use LLMs for planning. For instance, in robotics and navigation, the SayCan framework \cite{ahn2022icanisay} demonstrated the capability of LLMs to guide robots using natural language to do feasible and contextually appropriate actions. In planning, hybrid approaches have leveraged LLMs to generate subgoals for classical planners \cite{meng2025llmalargelanguagemodel,liu2023llm+,dagan2023dynamic}, while other methods integrate solver heuristics to better guide LLM generated plans \cite{wu2024llmsplanpathsextra,hirsch2024s}. Researchers have also explored combining LLMs with reinforcement learning (RL) to enhance reasoning capabilities. For instance, S2RCQL \cite{deng2025llmgoodpathplanner} augments prompts using information derived through Q-learning, and another approach employs LLMs to generate semantic hints for contextual RL in motion planning \cite{chen2025hcrmpllmhintedcontextualreinforcement}. Closer to our setting, recent work has brought LLMs directly into unmanned aerial vehicle (UAV) control: \emph{TypeFly} \cite{10970379} translates natural language into executable drone missions, and \emph{GSCE} \cite{wang2025gsce} introduces a structured prompt framework that improves the reliability of LLM-issued UAV commands. 

Moreover, LLMs have also shown strong potential as proxies for human preference, which can be utilized to readily evaluate preference alignment \cite{zheng2023judging}, and recent work on \emph{introspective planning} \cite{liang2024introspective} shows that aligning an LLM's uncertainty with task ambiguity yields safer downstream plans---motivating the coach/verifier agent we adopt in this work. A second complementary direction is retrieval-augmented planning, where an embodied agent conditions on trajectories or experiences retrieved from a memory store: P-RAG \cite{xu2024prag} introduces progressive retrieval for everyday planning tasks, and STRAP \cite{memmel2024strap} retrieves sub-trajectories for robot policy learning. FRAMe adopts the same intuition for eVTOL route planning, retrieving preference-conditioned past flight plans rather than learning a policy from scratch.

Although LLMs have been explored extensively across these domains, their potential for real-world flight planning remains largely unexplored. To address this gap, prior work introduced an approach in which an LLM selected the route most aligned with human preference from multiple candidates \cite{10749512}. That line of research was subsequently extended to an end-to-end flight planning generation system for eVTOL operations, employing various different chain-of-thought (CoT) prompting strategies \cite{DBLP:journals/corr/abs-2201-11903,tabrizian_chain--thought_2025}.

In this paper, we continue this progression by presenting an end-to-end LLM-based flight planning tool that incorporates a multi-modal coach agent and a memory system based on RAG. Specifically, we introduce FRAMe, a flight planning tool that enables human operators to generate eVTOL flight plans based on natural language preferences.

We propose a flight planning system that integrates a planner LLM with a multimodal coach agent and a RAG memory module. The LLM serves as the route planner, transforming textual prompts into sequences of geographic waypoints. The RAG module grounds the LLM's reasoning by supplying relevant context from prior planning experiences and their evaluations. The coach agent then validates and assesses each generated plan through a three-stage review: first, geometric tools verify that the flight plan is physically valid; second, the agent evaluates alignment with operator preferences by inspecting the rendered flight plan image; and third, the operator can optionally provide additional feedback. The resulting record is stored in the memory database, progressively improving future RAG retrievals. Our key contributions are as follows:
\begin{enumerate}
      \item \textbf{FRAMe System:} We design a flight planning framework that integrates a planner LLM with a RAG-based memory module and a multi-modal coach agent,
  enabling natural language-driven route generation that respects no-fly zones and aligns with operator preferences.
      \item \textbf{Annotator-Free Preference Evaluation:} We propose a quantitative evaluation framework built on three measurable preference objectives: minimizing flight distance, minimizing waypoint count, and maximizing polygon clearance. Because each objective is computed geometrically, preference alignment can be assessed objectively without human annotators. We pair this with a two-phase protocol, a warmup phase that seeds the memory followed by a read-only ablation phase, which isolates the contribution of retrieval augmentation.
      \item \textbf{Multi-Model Ablation:} We conduct a systematic evaluation across four LLMs (OpenAI's o3-mini, o4-mini, GPT-5.4 \cite{openai_models_2024}, and DeepSeek-R1 \cite{Guo_2025}) and three scenario difficulty levels, benchmarking against a classical A* baseline. The ablation isolates each component: the coach provides a non-redundant validity lift on top of retrieval for every planner---most visibly on o3-mini, where RAG alone does not improve over Baseline.
  \end{enumerate}
The remainder of this paper is as follows. Section~\ref{sect:sys_architecture} describes the FRAMe system architecture, including the planner agent, multi-modal coach agent, and RAG-based memory module. Section~\ref{sect:web} presents the web prototype and experimental evaluation, covering plan validity, preference-metric alignment, and comparison with an A* baseline across four LLMs. Section~\ref{sect:limitations} discusses limitations and Section~\ref{sect:conclusion} concludes. The appendix provides prompting strategies (Appendix~\ref{app:prompting_strategies}), the full coach vision agent prompt (Appendix~\ref{app:coach_prompt}), qualitative flight-plan examples (Appendix~\ref{app:qualitative_extended}), a per-preference metric breakdown (Appendix~\ref{app:pref_matrix}), and a per-difficulty validity analysis (Appendix~\ref{app:validity_difficulty}).

\section{System Architecture} \label{sect:sys_architecture}

\begin{figure*}[!h]
  \centering
  \includegraphics[width=\linewidth]{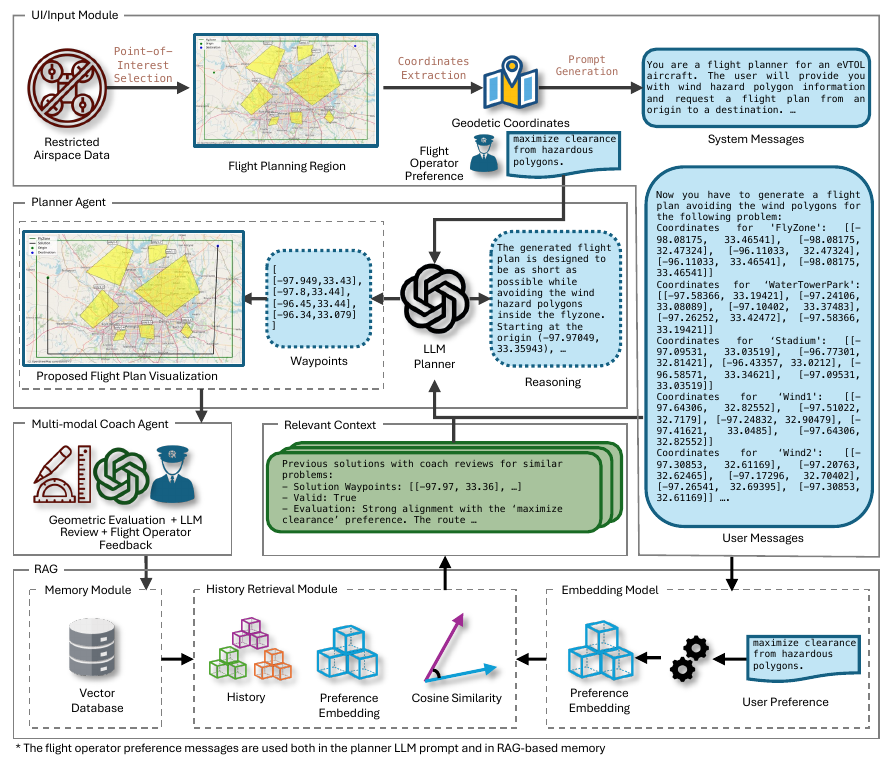}
  \caption{FRAMe System Architecture: The planner LLM generates a flight plan from a natural language prompt and scenario data. A RAG module provides relevant context to the LLM, such as previous similar planning experiences and their evaluations. The multi-modal LLM coach agent evaluates the planner’s proposed route for safety and compliance with the preference. Then, it supplies review that will be combined with the flight operator's feedback and will be stored in a database.}
  \label{fig:arch}
\end{figure*}

FRAMe’s architecture combines an LLM-based planning core with auxiliary modules for solution evaluation and knowledge retrieval. The system takes two main inputs: a flight scenario comprised of an origin, a destination, the flyable airspace, and a set of no‑fly zones encoded as polygons in a KML file, and a natural language prompt from the flight operator describing how the route should traverse the environment (for example, ``Maximize clearance from hazardous polygons.''). KML is a file format that can visualize geographic data in earth browsers such as Google Earth \cite{google2023_kmltutorial}. The flowchart of the system architecture is depicted in Figure \ref{fig:arch}. We explain FRAMe components here.
\subsection{Planner Agent}
The planning process begins by extracting the origin, destination, flyable airspace, and no‑fly‑zone placemarks from the operator’s KML file and converting them into natural‑language messages. The system messages are generated according to the chosen prompting strategy, priming the LLM with general instructions. Tabrizian et al. \cite{tabrizian_chain--thought_2025} discussed different prompting strategies based on CoT and showed that the best results are achieved by using a customized version of CoT without providing any examples (Zero-shot). All of the prompting strategies available in FRAMe are described in the Description of Different Prompting Strategies section in Appendix. After system messages are constructed, they will be combined with user messages (natural-language information of the KML) and the flight operator preferences. This will be the input for the LLM-planner. If RAG is enabled, these messages will be augmented with prior successful flight plans based on the flight setting and operator preferences. With the fully assembled prompt, the LLM-planner produces a candidate flight plan which is a sequence of waypoints forming a flight path that begins at the origin, ends at the destination, avoids all no‑fly zones, and adheres as closely as possible to the operator’s preference. The planner will also generate a reasoning in natural language for better explainability purposes.
\subsection{Multi-modal Coach Agent}
The multimodal coach agent is implemented using o4-mini alongside rule-based geometric checks and flight operator's feedback. The agent first verifies plan validity. 
A plan is considered valid if:
(1) All generated waypoints remain within the flyable airspace, (2) the plan correctly connects the designated origin and destination based on the scenario data, and (3) no segment passes through a no-fly zone.

\begin{figure}[!h]
  \centering
  \includegraphics[width=0.9\linewidth]{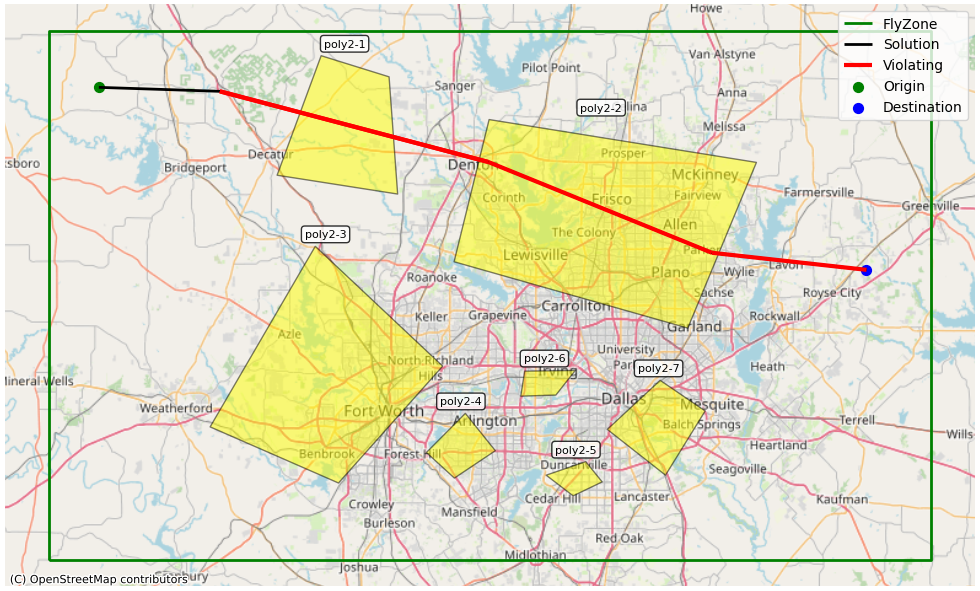}
  \caption{A sample of the generated image of a flight plan for the multi-modal coach agent to review.}
  \label{fig:img_flightplan}
\end{figure}

All criteria are evaluated using geometric tools. Following the validity check, the system renders an image of the full planning problem with the proposed solution overlaid (see Figure~\ref{fig:img_flightplan}) and assesses preference alignment, confirming whether the generated route satisfies the operator's directional or positional instructions using o4-mini's vision capabilities (for complete coach vision agent prompt see Appendix \ref{app:coach_prompt}). The vision agent has access to both the rendered solution image and the validity evaluation results. Finally, the flight operator can optionally provide free-form feedback on the proposed plan. 

\subsection{Retrieval Augmented Generation}
To augment the prompt with proper context about the current flight planning problem, we implement the RAG module. When enabled, the system queries a vector database to retrieve past flight plans that most closely resemble the current setting and the operator’s stated preferences.

The process begins by passing the operator’s preferences into an embedding model, which converts the textual input into a numerical embedding vector. This vector is then used by the history retrieval module to identify the most similar prior flight plans based on the cosine distance between the current preference embedding $\mathbf{e_p}$ and each other preference embedding in the database $\forall \mathbf{e_h} \in \mathcal{H}$:
$$
\text{CosineDistance}(\mathbf{e_p}, \mathbf{e_h}) = 1 - \frac{\mathbf{e_p} \cdot \mathbf{e_h}}{\|\mathbf{e_p}\| \|\mathbf{e_h}\|},
$$
 where $\mathcal{H}$ denotes the set of preference embedding vectors corresponding to previously successful flight plans. Since lower cosine distances indicate greater similarity, the system selects records with the $K$ lowest values. Note that the retrieval query restricts the candidate set to prior records with identical scenario geometry, that is, the same flyzone, no-fly polygons, origin, and destination. The cosine similarity over the operator's natural-language preference text is then used only to rank within that geometrically identical set. Geometric relevance is thus guaranteed by exact scenario matching rather than inferred from the embedding distance, and the preference embedding only selects which past plan for the same problem best matches the current intent. The retrieved plans, referred to as the contextual history, represent flight plans most aligned with the current setting and preferences.

The contextual history is then used to augment the input prompt, providing relevant context and guidance for generating the new flight plan. The planner LLM processes this enriched prompt to produce a flight plan, which is presented to the operator for evaluation. If the plan is submitted for storage, the RAG module then collects the relevant data into the vector database, making it available for future retrieval. The complete planning workflow is detailed in Algorithm \ref{algorithm}.

\begin{algorithm}[t]
\small
\caption{FRAMe: Multi-modal Coach Agent and RAG}
\label{algorithm}
\begin{algorithmic}[1]
  \REQUIRE Operator preference $P$, prompting strategy $S$, database $\mathcal{DB}$, module flags \textit{RAG} and \textit{Coach}, retrieval count $K$, scenario $C$ with origin $O$, destination $D$, and no-fly zones $Z=\{z_i\}_{i=1}^n$
  \ENSURE Waypoints $W=\{w_j\}_{j=1}^{m}$ and reasoning $R$
  \STATE \textbf{$\triangleright$ Step 1: Build prompt}
  \STATE $\textit{sc} \gets \text{GetScenario}(C, O, D, Z)$
  \STATE $M_{\text{sys}} \leftarrow \text{SysMsg}(S)$
  \STATE $M_{\text{user}} \leftarrow \text{UsrMsg}(P,\, \textit{sc})$
  \STATE $\textit{prompt} \leftarrow M_{\text{sys}} + M_{\text{user}}$
  \STATE $e_p \gets \text{Embed}(P)$

  \IF{\textit{RAG}}
    \STATE $\textit{cands} \leftarrow \text{Query}(\mathcal{DB},\, \textit{sc})$
    \STATE $\{e_h\}_{h \in \mathcal{H}} \gets \text{EmbedAll}(\textit{cands})$
    \STATE $\textit{dists} \gets \{\text{CosDist}(e_h, e_p)\}_{h \in \mathcal{H}}$
    \STATE $\textit{hist} \leftarrow \text{TopK}(K,\, \textit{cands},\, \textit{dists})$
    \STATE $\textit{ctx} \leftarrow \text{GetContext}(\textit{hist})$
    \STATE $\textit{prompt} \gets \textit{prompt} + \textit{ctx}$
  \ENDIF

  \STATE $\textit{resp} \leftarrow \text{PlannerLLM}(\textit{prompt})$
  \STATE $(W, R) \leftarrow \text{ParsePlan}(\textit{resp})$

  \IF{\textit{Coach}}
    \STATE \textbf{$\triangleright$ Step 2: Validity and alignment checks}
    \STATE $\textit{val} \leftarrow \text{Coach.EvalValidity}(W,\, \textit{sc})$
    \STATE $\textit{aln} \leftarrow \text{Coach.EvalAlignment}(W,\, \textit{sc},\text{SolutionImg})$

    \STATE \textbf{$\triangleright$ Step 3: Operator feedback}
    \STATE $\textit{fb} \gets \text{GetFeedback}(W, R,\, \textit{val},\, \textit{aln},\, \textit{sc})$

    \STATE \textbf{$\triangleright$ Step 4: Update database}
    \STATE $\textit{rec} \leftarrow \{\textit{sc},\; e_p,\; \textit{val},\; \textit{aln},\; \textit{fb},\; W,\; R\}$
    \STATE $\text{Insert}(\mathcal{DB},\;\textit{rec})$
  \ENDIF

  \STATE \textbf{return} $W$, $R$
\end{algorithmic}
\end{algorithm}
\section{Web Prototype and User Interface} \label{sect:web}
Conducting real flight tests with full-scale eVTOL aircraft is prohibitively expensive at this stage of research. Although FRAMe is designed and evaluated in the context of eVTOL mission planning, we deploy it as a web application targeting subscale UAV operations as a practical intermediate step. The web interface connects with small UAVs and ground control station software such as Mission Planner \cite{ArduPilotMissionPlannerOverview}, a widely used platform for configuring, controlling, and monitoring UAVs. This is intended to allow validation of the planning pipeline in real flight conditions at a fraction of the cost, while the underlying planning logic and preference-alignment framework remain directly applicable to eVTOL operations. The backend is implemented in Python, integrating the LLM via an API and our custom modules for retrieval and validation. The front-end is a user-friendly interface that guides the operator through scenario setup, preference input, and reviewing the flight plan. 
\begin{figure}[!ht]
  \centering
  \includegraphics[width=0.95\linewidth]{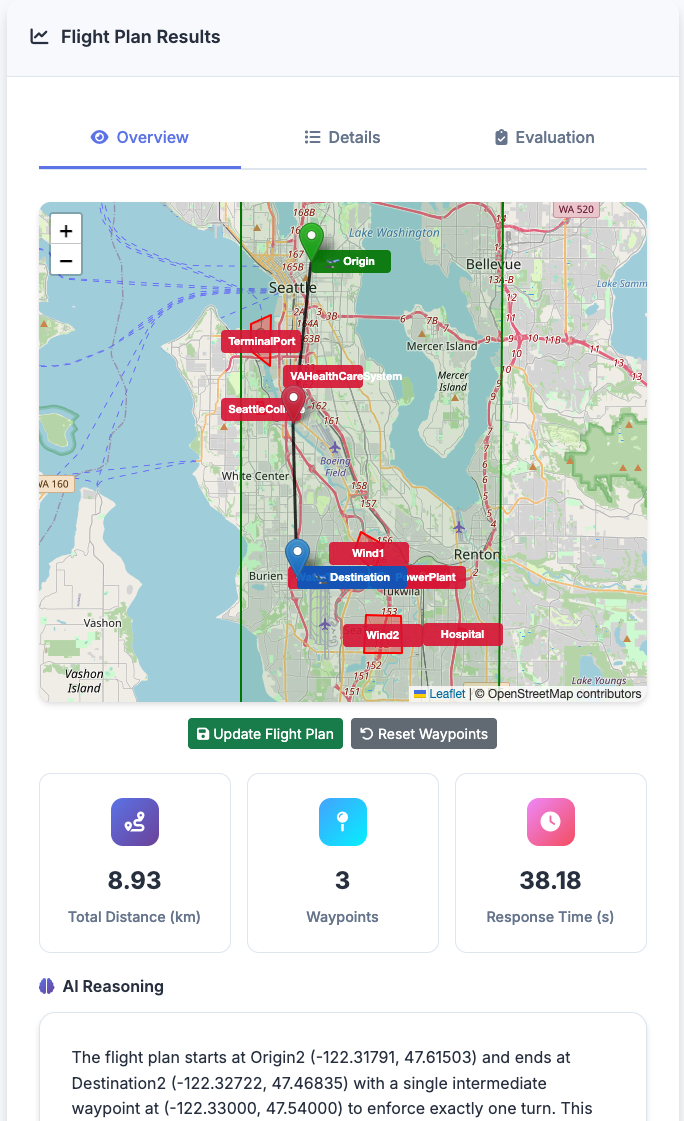}
  \caption{Overview tab in the web user interface: a leaflet map shows the computed route (black) with origin (green), destination (blue), and no‐fly zone boundaries (red). AI-reasoning will be provided in this tab.}
  \label{fig:results}
\end{figure}

\begin{figure*}[!h]
  \centering
  \includegraphics[width=0.8\linewidth]{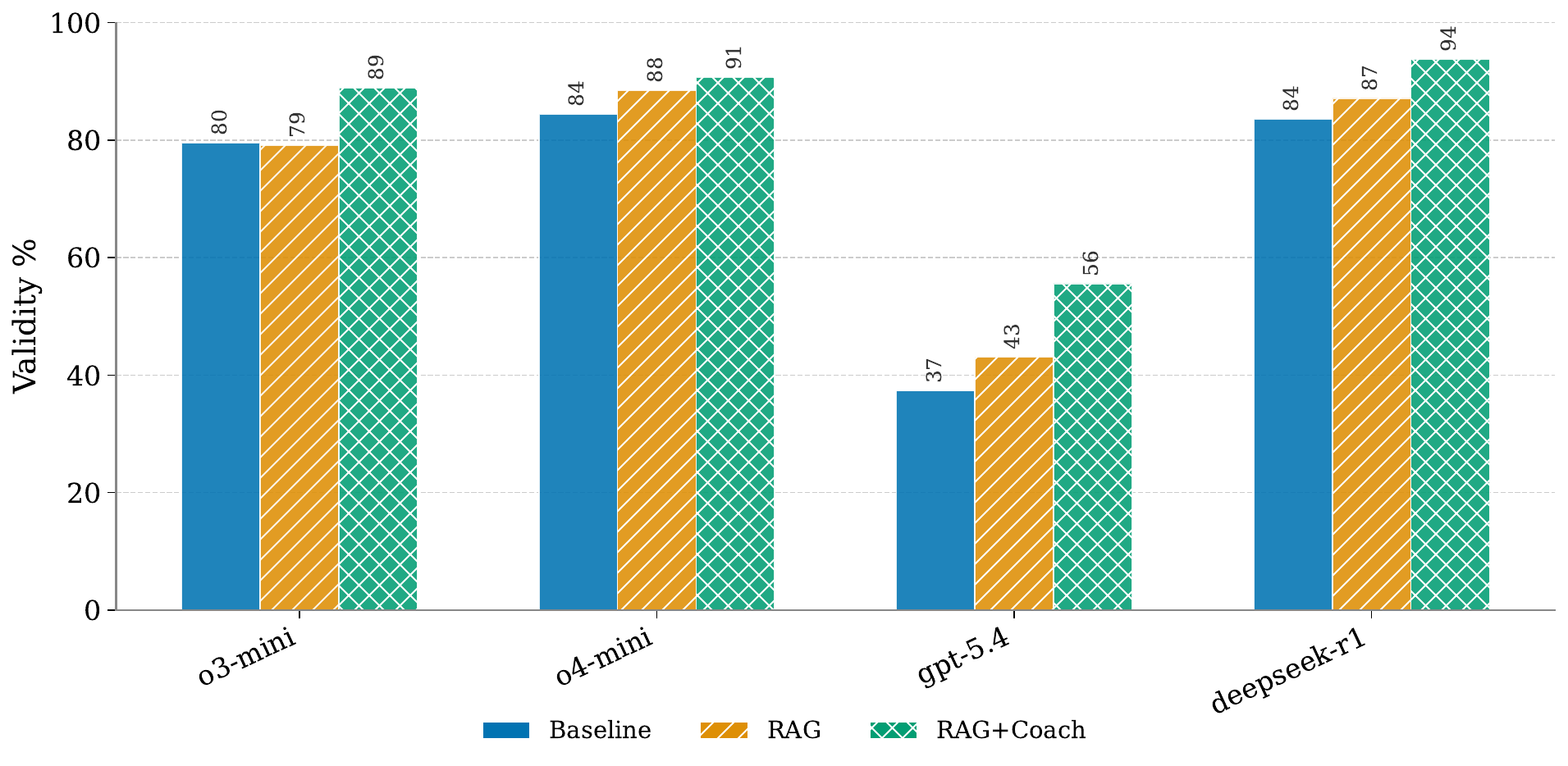}
  \caption{Validity rate (\%) by condition for all planners. Conditions are Baseline, +RAG, and +RAG+Coach.}
  \label{fig:validity_barchart}
\end{figure*}

\begin{figure*}[!h]
  \centering
  \includegraphics[width=\linewidth]{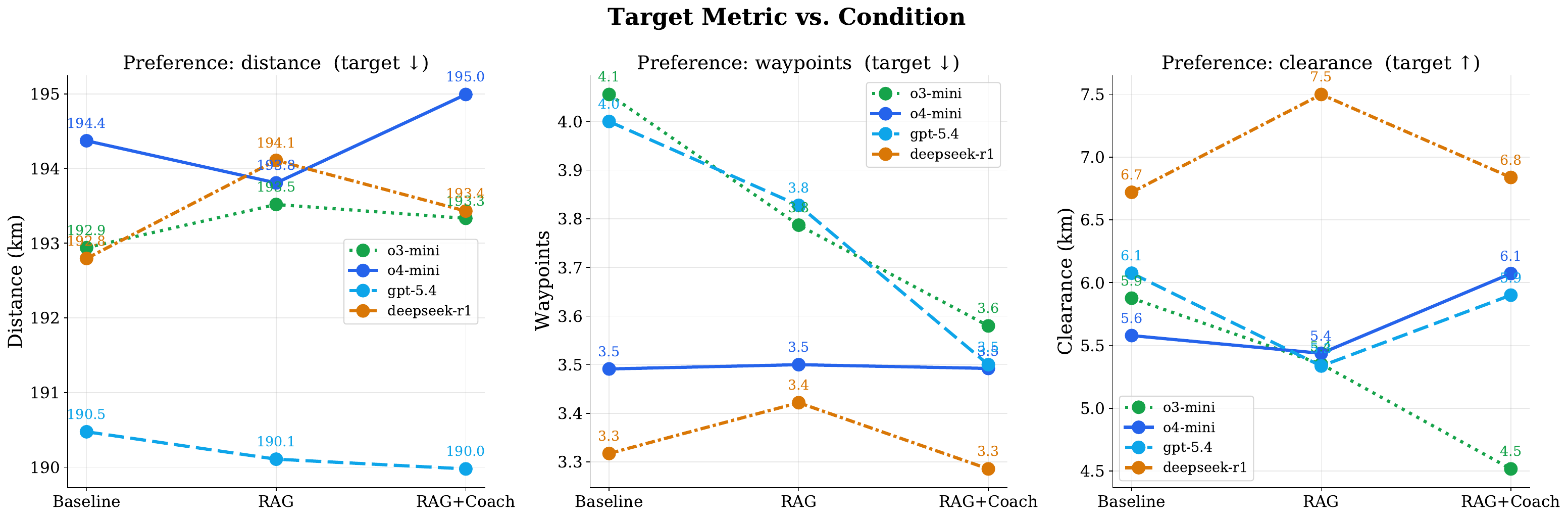}
  \caption{Mean of the preference-relevant metric versus condition for each of the three preferences (valid plans only). Desired direction marked above each panel.}
  \label{fig:alignment_lines}
\end{figure*}

\subsection{Experimental Setup}

We evaluate FRAMe on flight scenarios in the Dallas--Fort Worth metropolitan area grouped into three difficulty levels---\emph{Easy}, \emph{Medium}, and \emph{Hard}---that differ in the number and geometry of no-fly zones (2, 4, and 7 polygonal restrictions, respectively). Each difficulty level is paired with every combination of a fixed set of origins and destinations, producing a broad coverage of flight configurations.

\textbf{Models.} We benchmark four state-of-the-art LLMs as the planning module: \emph{OpenAI o3-mini}, \emph{OpenAI o4-mini}, \emph{DeepSeek-R1}, and \emph{OpenAI GPT-5.4}.

\textbf{Conditions.} Each scenario is evaluated under four conditions that reflect progressively richer augmentation. We use the same short names (\textbf{A*}, \textbf{Baseline}, \textbf{+RAG}, \textbf{+RAG+Coach}) throughout the text, tables, and figures:
\begin{enumerate}
    \item \textbf{A*}: a classical A* path planner serving as a geometry-optimal baseline.
    \item \textbf{Baseline}: the LLM planner alone, with no retrieval augmentation or coach feedback.
    \item \textbf{+RAG}: the LLM planner augmented with the RAG module, retrieving the $K{=}2$ most relevant prior plans from the database (no coach).
    \item \textbf{+RAG+Coach}: the full FRAMe system, combining RAG-based retrieval with the multi-modal coach agent for validity checking and preference verification.
\end{enumerate}   
\textbf{Preferences.} To enable objective, quantitative evaluation of preference alignment, we use three measurable preference objectives: (1) \emph{minimize total flight distance}, (2) \emph{minimize the number of waypoints}, and (3) \emph{maximize clearance from hazardous polygons}. For each preference, alignment is assessed by whether the model’s output improves the corresponding metric relative to the other preference conditions within the same scenario.

\textbf{Warmup and ablation protocol.} Before ablation, a warmup phase runs each planner over every scenario with a neutral preference prompt, both with and without the coach agent, retaining only the plans that pass the geometric validity checks, to seed the RAG vector database (150 runs). The ablation phase then fixes the database in read-only mode and sweeps each model across the full scenario--preference--condition grid, isolating the marginal contribution of RAG and of the coach.

\subsection{Results}

\textbf{Plan validity.} Figure~\ref{fig:validity_barchart} reports the validity rate for all four models averaged over difficulty levels, and Table~\ref{tab:reasoning_main} provides the corresponding preference-metric breakdown for valid plans. For every model, +RAG+Coach achieves the highest validity. The three reasoning models converge to 88.9--93.8\% under the full system (o3-mini: 79.6\%$\rightarrow$79.1\%$\rightarrow$88.9\%, o4-mini: 84.4\%$\rightarrow$88.4\%$\rightarrow$90.7\%, DeepSeek-R1: 83.6\%$\rightarrow$87.1\%$\rightarrow$93.8\%), while GPT-5.4 lags considerably, improving from a markedly lower Baseline of 37.3\% to 55.6\%---a gap that persists across all conditions and reflects weaker instruction-following on geometrically constrained scenarios. For \emph{o3-mini} in particular, +RAG alone slightly regresses validity (79.6\%$\rightarrow$79.1\%)---retrieval without validation surfaces misleading neighbors---and the coach is what recovers and extends the gain, indicating that the coach review is load-bearing rather than a redundant signal on top of retrieval.

\begin{table}[t]
\centering
\scriptsize
\setlength{\tabcolsep}{2pt}
\caption{Preference-metric results (valid plans only, averaged over all difficulties and preferences). $\Delta$ is relative to A*: positive $\Delta$ is favorable for Clr.\ but unfavorable for Dist.\ (LLMs fly farther than A*) and \#WP. Best LLM value per column in \textbf{bold}.}
\label{tab:reasoning_main}
\begin{tabular}{@{}llccc@{}}
\toprule
Model & Cond. & Dist.\,(km, $\Delta$) & Clr.\,(km, $\Delta$) & \#WP\,($\Delta$) \\
\midrule
A* & --- & 189.9 & 1.91 & 18.2 \\
\midrule
\multirow{3}{*}{o3-mini}
 & Baseline   & 195.5 ($+$5.6) & 4.65 ($+$2.74) & 5.17 ($-$13.0) \\
 & +RAG       & 194.7 ($+$4.8) & 4.36 ($+$2.45) & 4.65 ($-$13.5) \\
 & +RAG+Coach & 194.5 ($+$4.6) & 3.55 ($+$1.64) & 4.42 ($-$13.8) \\
\midrule
\multirow{3}{*}{o4-mini}
 & Baseline   & 195.2 ($+$5.3) & 3.90 ($+$1.99) & 3.94 ($-$14.3) \\
 & +RAG       & 196.1 ($+$6.2) & 4.03 ($+$2.12) & 3.81 ($-$14.4) \\
 & +RAG+Coach & 196.6 ($+$6.7) & 4.36 ($+$2.45) & \textbf{3.75 ($-$14.4)} \\
\midrule
\multirow{3}{*}{GPT-5.4}
 & Baseline   & \textbf{190.5 ($+$0.6)} & 4.98 ($+$3.07) & 5.36 ($-$12.8) \\
 & +RAG       & 191.7 ($+$1.8) & 4.43 ($+$2.52) & 4.68 ($-$13.5) \\
 & +RAG+Coach & 191.2 ($+$1.3) & 4.20 ($+$2.29) & 4.23 ($-$14.0) \\
\midrule
\multirow{3}{*}{DeepSeek-R1}
 & Baseline   & 195.7 ($+$5.8) & 4.75 ($+$2.84) & 3.80 ($-$14.4) \\
 & +RAG       & 196.6 ($+$6.7) & \textbf{5.25 ($+$3.34)} & 3.83 ($-$14.4) \\
 & +RAG+Coach & 195.9 ($+$6.0) & 5.01 ($+$3.10) & \textbf{3.75 ($-$14.4)} \\
\bottomrule
\end{tabular}
\end{table}

\textbf{Preference capture.} Figure~\ref{fig:alignment_lines} traces how the preference-relevant metric evolves across conditions for valid plans, with reasoning models showing distinct preference-specific patterns. Under the \emph{waypoints} preference, o3-mini is the strongest reasoning-model responder, reducing mean waypoint count from 4.06 (Baseline) to 3.58 (+RAG+Coach, $-12\%$); o4-mini and DeepSeek-R1 start already near their planning floor (3.49 and 3.32, respectively), leaving little room for further reduction. Under the \emph{clearance} preference, o4-mini shows the clearest gain, increasing minimum clearance from 5.58\,km to 6.07\,km ($+9\%$); DeepSeek-R1 peaks at +RAG (7.50\,km, up from 6.72\,km at Baseline) but retreats slightly with the coach (6.84\,km), suggesting that retrieval alone is the stronger signal for its spatial reasoning. Notably, o3-mini's clearance declines monotonically with augmentation (5.88$\rightarrow$5.35$\rightarrow$4.52\,km in Figure~\ref{fig:alignment_lines}; the per-preference matrix in Appendix~\ref{app:pref_matrix} reports the same downward trend with rounding to one decimal)---coach-driven revisions appear to prioritize correcting validity violations over maximizing polygon separation. The \emph{distance} preference moves little for all three reasoning models (193--197\,km), reflecting that they already operate close to their planning floor at Baseline. GPT-5.4, by contrast, is the only model that monotonically improves distance \emph{when the distance preference is active} (190.5$\rightarrow$190.0\,km in Figure~\ref{fig:alignment_lines}); we note this is a per-preference effect---in the all-preferences aggregate of Table~\ref{tab:reasoning_main} GPT-5.4's mean distance rises slightly with augmentation (190.5$\rightarrow$191.7$\rightarrow$191.2\,km), so the gain is conditional on the operator actually requesting distance. GPT-5.4 also shows comparable waypoint reduction ($-13\%$), though its lower overall validity limits the practical impact of these gains.

\textbf{A* baseline comparison.} The classical A* planner achieves 100\% validity by construction, but it is preference-blind: its mean clearance of 1.91\,km is well below what every LLM produces even in the Baseline condition (3.9--5.0\,km), and its waypoint count (18.2) is $3$--$5\times$ higher than every LLM configuration because A* tracks the grid discretization rather than the operator's intent. FRAMe (even without RAG) already outperforms A* on the clearance and waypoint preferences, while accepting a small distance overhead over the preference-blind geodesic. Adding +RAG+Coach widens the clearance and waypoint margins and recovers a substantial part of the validity gap for the stronger planners, most clearly for DeepSeek-R1.

\section{Limitations} \label{sect:limitations}

\textbf{Headroom-bounded preference capture.} The distance preference does not move meaningfully under any condition (Table~\ref{tab:reasoning_main}) because every model already plans within 6\,km of the $\approx$190\,km geodesic at Baseline. The same headroom caveat partially applies to waypoint count for o4-mini and DeepSeek-R1, which sit near a 3.3--3.5 floor at Baseline and therefore have limited room to improve further. Our preference-capture evidence is therefore strongest for the clearance preference (and for waypoint count on planners that start above the floor, e.g.\ o3-mini and GPT-5.4) and weakest for distance---a consequence of the evaluation geometry rather than evidence that the framework ignores the preference.

\textbf{Aggregate vs.\ per-preference effects.} Several of our positive findings are clearest when conditioned on the matching preference (Figure~\ref{fig:alignment_lines}). When metrics are averaged across all three preferences (Table~\ref{tab:reasoning_main}), augmentation can move a non-target metric in the unfavorable direction---for example, aggregate clearance for o3-mini and GPT-5.4 declines under +RAG+Coach as those configurations correctly de-prioritize clearance when the operator did not request it. The aggregate column should therefore be read as a side-effect summary, not as the primary preference-capture signal.

\textbf{LLM-as-judge bias in the coach.} The preference-alignment step of the coach is performed by a multimodal LLM (o4-mini) evaluating the planner's output, and prior work has shown that LLM judges can exhibit position, verbosity, and self-preference biases \cite{zheng2023judging}. We mitigate this by combining the vision judgment with rule-based geometric checks for validity, but the alignment verdict itself inherits the hidden biases or priors the judge carries. A human-rater study on a subset of plans is left for future work.

\textbf{Simulation-only evaluation.} All experiments use synthetic polygonal no-fly zones; no live flight tests or real weather data are incorporated. As a next step, we plan to validate the approach through real flight tests with a subscale drone.

\textbf{Constraint coverage.} Our current formulation treats hazards as static no-fly polygons and does not model dynamic traffic from other UAVs or crewed aircraft, onboard energy or battery limits, or vehicle dynamics. These factors are safety critical for feasibility and collision avoidance in dense airspace, and incorporating them, for example by adding traffic and energy terms to the scenario and the validity checks, is important future work.

\section{Conclusion} \label{sect:conclusion}
We presented FRAMe, an end-to-end flight planner that couples an LLM with a RAG memory of prior plans and a multi-modal coach agent that gates both validity and preference alignment. The framework uses geometric ground truth, so it needs no human annotators, and it is modular: retrieval and the coach can be ablated independently against a fixed warmup-seeded database. Across four state-of-the-art LLMs and three operator preferences, the coach review is load-bearing on top of retrieval, attaining the highest validity for every planner; for the weakest planner, retrieval alone can slightly regress validity, and the coach is what recovers it. Preference capture is real but model-specific: planners with headroom shift the requested metric in the operator-favored direction, while those already near their planning floor have little room to move, and gains can surface under retrieval alone or only once the coach is added. Finally, FRAMe is preference-aligned where A* is preference-blind, accepting a small distance overhead in exchange for higher clearance and substantially fewer waypoints.

\bibliography{example_paper}
\bibliographystyle{icml2026}

\newpage
\appendix
\onecolumn
\section{Description of Different Prompting Strategies} \label{app:prompting_strategies}
Here we describe all of the prompting strategies available for FRAMe:
\paragraph{Raw Prompt:} This baseline uses minimal prompt engineering. The scenario and request are provided directly to the LLM without any additional guidance on the solution approach. Specifically, the system message is:
\begin{promptbox}{Raw Prompt}
``You are a flight planner for an eVTOL aircraft. The user will provide you with wind hazard polygon information and request a flight plan from an origin to a destination. You must generate a flight plan as a list of waypoints starting from the origin coordinate and ending at the destination coordinate while avoiding the wind polygons. Always include both the origin and destination points in your response. You can generate as many waypoints as necessary to avoid the polygons. More waypoints may lead to a smoother flight plan. You cannot fly outside of the fly zone.''
\end{promptbox}
There is no task breakdown or few-shot example. This represents how the model might perform ``out of the box'' given a straightforward request.

\paragraph{Zero-Shot Prompt (Basic):} In the zero-shot setting, we append the phrase “Think step by step.” to the system messages, following \cite{DBLP:journals/corr/abs-2201-11903}, to prompt a strategic reasoning process. However, we do not specify the individual steps the LLM should follow for planning, nor do we supply any examples in this approach.

\paragraph{Zero-Shot Prompt (Customized):} This approach is a modified form of zero-shot prompting that provides more explicit guidance for reasoning. The system message is extended with instructions tailored for chain-of-thought reasoning in flight planning tasks. The following steps are incorporated into the system messages:
\begin{promptbox}{Zero-Shot Prompt (Customized)}
``You are a flight planner for an eVTOL aircraft. The user will provide you with hazardous polygon information and request a flight plan from an origin to a destination. You must generate a flight plan as a list of waypoints starting from the origin coordinate and ending at the destination coordinate while avoiding the hazardous polygons. Always include both the origin and destination points in your response. You can generate as many waypoints as necessary to avoid the polygons---more waypoints may lead to a smoother flight plan. You cannot fly outside of the flyzone. The best approach to find the optimal solution is as follows: (1) Identify the origin and destination points. (2) Identify the hazardous polygons and the flyzone. (3) IMPORTANT STEP: Generate waypoints that connect the origin to the destination while avoiding hazardous polygons and staying within the flyzone (they should not be on the flyzone's border either). You may generate more waypoints near the hazardous polygons to ensure that the line segments do not intersect with the hazardous polygons. Ensure that the flight plan connecting the waypoints are aligned with the human preference. YOU NEED TO INCLUDE AT LEAST 4 DECIMAL POINTS FOR WAYPOINT COORDINATES. (4) The line segments connecting the waypoints should not have sharp angles (recommended). (5) Ensure that the line segments do not intersect with the hazardous polygons. (6) If any of the line segments intersect with the polygons, modify the corresponding waypoints so the new line segment does not intersect the polygon.''
\end{promptbox}
By clarifying the criteria and encouraging intermediate reasoning, this strategy is expected to help the LLM perform better. However, no example flight plan is given, so this remains a zero-shot approach (the model must draw on its internal knowledge and the on-the-fly reasoning).

\paragraph{One-Shot Prompt (Easy Example):}
This method resembles the zero-shot (customized) approach, but additionally supplies the LLM with a single example problem–solution pair before presenting the new scenario. The example is a simple case (structurally similar but with different coordinates) that illustrates the format of a correct solution.

\paragraph{One-Shot Prompt (Hard Example):}
This approach is analogous to the previous one, except that the provided example depicts a challenging scenario with multiple hazards, illustrating a more elaborate chain-of-thought process.

\section{Coach Vision Agent Prompt}\label{app:coach_prompt}

The multi-modal coach agent receives two inputs simultaneously: (i) a structured text prompt, shown below, in which \texttt{\{geometric\_summary\}} is filled at runtime with the output of the rule-based geometric checks (validity flag, violated polygons, waypoints outside the flyzone, and origin/destination match), and \texttt{\{human\_preference\}} with the operator's stated preference; and (ii) the rendered flight plan image depicted in Figure~\ref{fig:img_flightplan}. The agent first handles invalid plans (Step~0), then---only if the plan is geometrically valid---judges preference alignment from the image (Step~1), returning a structured verdict (\texttt{aligned}, \texttt{evaluation}, \texttt{reasoning}).

\begin{promptbox}{Coach Vision Agent Prompt}
You are a senior flight-operations reviewer evaluating an eVTOL flight plan that was produced by an automated planner.

\smallskip
\textbf{SCOPE}\\
Validity checks (polygon intersection, flyzone containment, origin/destination endpoints) have already been performed by geometric tools upstream. DO NOT re-evaluate validity. Your ONLY job is to judge how well the flight plan aligns with the flight operator's preference, or---if no preference is given---how optimal the plan looks.

\smallskip
\textbf{LEGEND} (what you will see in the image)
\begin{itemize}[noitemsep,topsep=2pt]
  \item Green rectangle: the flyzone.
  \item Yellow polygons (labeled poly1-1, poly1-2, \ldots): hazardous zones.
  \item Green dot labeled ``Origin'': start of the route.
  \item Blue dot labeled ``Destination'': end of the route.
  \item Black line segments: path segments not intersecting any hazardous polygon.
  \item Red line segments: path segments intersecting a hazardous polygon. (Informational only---ignore for this review.)
\end{itemize}

\smallskip
\textbf{GEOMETRIC EVALUATION} (ground truth from upstream tools---trust this over the image)\\
\texttt{\{geometric\_summary\}}

\smallskip
\textbf{FLIGHT OPERATOR PREFERENCE}\\
\texttt{\{human\_preference\}}

\smallskip
\textbf{EVALUATION TASK}

\textbf{STEP 0} --- Check validity first (from the GEOMETRIC EVALUATION block above):
\begin{itemize}[noitemsep,topsep=2pt]
  \item If ``Valid overall'' is False, the path is INVALID. Do NOT judge preference alignment or optimality. Return: \texttt{aligned = False}, \texttt{evaluation = ``Path is invalid---alignment not evaluated.''}, \texttt{reasoning} = a short summary of which geometric rule was broken. Stop here.
\end{itemize}

\textbf{STEP 1} --- Only if the path is valid, evaluate alignment:
\begin{itemize}[noitemsep,topsep=2pt]
  \item If a meaningful preference is provided: judge how well the path follows that preference. Be specific about which part of the route satisfies or violates it. Set \texttt{aligned = True} only if the path clearly follows the preference.
  \item If no meaningful preference is provided: judge the path on optimality only---is it close to the shortest reasonable route? Are there unnecessary detours, zig-zags, or sharp turns? Set \texttt{aligned = True} if the path is near-optimal, False otherwise.
\end{itemize}

Use the geometric evaluation above as hard facts. For the alignment judgment, focus on the VISUAL pattern: shape of the detour, which side of polygons the path goes around, spacing of waypoints, distance from hazards. Cite concrete visual evidence. Do not speculate about data you cannot see in the image.

\smallskip
\textbf{OUTPUT}\\
Return a structured object with: \texttt{aligned} (bool), \texttt{evaluation} (one- or two-sentence verdict).
\end{promptbox}

\section{Extended Qualitative Flight-Plan Examples}\label{app:qualitative_extended}

The examples below show the real inputs and outputs of two valid and one invalid FRAMe runs with the \emph{o4-mini} planner. Within each \emph{FRAMe Input} box, three sections are color-coded: \textcolor{syscolor}{\textbf{SYS}} (the fixed system message, shared across all runs), \textcolor{usercolor}{\textbf{USER}} (the per-scenario user message built by \texttt{prompt\_generator}), and \textcolor{operatorcolor}{\textbf{OPERATOR}} (the operator preference, appended to the user message). The \emph{FRAMe Response} box shows the waypoints and natural-language explanation returned by the planner. Finally, coach review is shown for the generated flight plan. Polygon labels such as poly7-1 and poly1-3 use the scenario identifier as a prefix; the leading number is a scenario ID, not the polygon count.

\subsection{Easy scenario, clearance preference}

\begin{inputbox}
{\color{syscolor}\textbf{SYS:}} You are a flight planner for an eVTOL aircraft. The user will provide you with hazardous polygon information and request a flight plan from an origin to a destination. You must generate a flight plan as a list of waypoints starting from the origin coordinate and ending at the destination coordinate while avoiding the hazardous polygons. Always include both the origin and destination points in your response. You can generate as many waypoints as necessary to avoid the polygons---more waypoints may lead to a smoother flight plan. You cannot fly outside of the flyzone. The best approach to find the optimal solution is as follows: (1) Identify the origin and destination points. (2) Identify the hazardous polygons and the flyzone. (3) IMPORTANT STEP: Generate waypoints that connect the origin to the destination while avoiding hazardous polygons and staying within the flyzone (they should not be on the flyzone's border either). You may generate more waypoints near the hazardous polygons to ensure that the line segments do not intersect with the hazardous polygons. Ensure that the flight plan connecting the waypoints are aligned with the human preference. YOU NEED TO INCLUDE AT LEAST 4 DECIMAL POINTS FOR WAYPOINT COORDINATES. (4) The line segments connecting the waypoints should not have sharp angles (recommended). (5) Ensure that the line segments do not intersect with the hazardous polygons. (6) If any of the line segments intersect with the polygons, modify the corresponding waypoints so the new line segment does not intersect the polygon.

\smallskip
{\color{usercolor}\textbf{USER:}} Now you have to generate a flight plan avoiding the hazardous polygons for the following problem. Coordinates for `FlyZone': (\mbox{$-$98.082},\,33.465), (\mbox{$-$98.082},\,32.473), (\mbox{$-$96.110},\,32.473), (\mbox{$-$96.110},\,33.465). Coordinates for `poly7-1': (\mbox{$-$97.319},\,33.093), (\mbox{$-$96.749},\,33.028), (\mbox{$-$96.667},\,33.248), (\mbox{$-$97.164},\,33.304). Coordinates for `poly7-2': (\mbox{$-$97.133},\,32.631), (\mbox{$-$96.754},\,32.587), (\mbox{$-$96.679},\,32.736), (\mbox{$-$96.964},\,32.852). Coordinates for `Origin2': (\mbox{$-$97.966},\,33.149). Coordinates for `Destination4': (\mbox{$-$96.255},\,32.826).

\smallskip
{\color{operatorcolor}\textbf{OPERATOR:}} ``Maximize clearance from hazardous polygons.''
\end{inputbox}

\begin{frameresponse}
\textbf{Waypoints:} (33.1490,\,\mbox{$-$97.9662}) $\to$ (33.4500,\,\mbox{$-$97.9662}) $\to$ (33.4500,\,\mbox{$-$96.2548}) $\to$ (32.8261,\,\mbox{$-$96.2548}).

\smallskip
\textbf{Reasoning:} ``To maximize clearance from the hazardous polygons, the flight plan goes north to the upper flyzone boundary ($33.45$°N), then east to the destination's longitude, then south to the destination. This path stays far from all polygons.''

\smallskip
\end{frameresponse}
\begin{coachreview}
\textbf{Valid:} True \quad \textbf{Waypoints outside flyzone:} \{\} \quad \textbf{Violated polygons:} {[}{]}

\smallskip
\textbf{Coach Verdict:} {[ALIGNED]}

\smallskip
\textbf{Evaluation:} Strong alignment with the `maximize clearance' preference. The route climbs to the northern border, traverses the top edge, and descends at the far east---staying well away from both hazardous polygons.
\end{coachreview}
\begin{figure}[!h]
  \centering
  \includegraphics[width=0.6\linewidth]{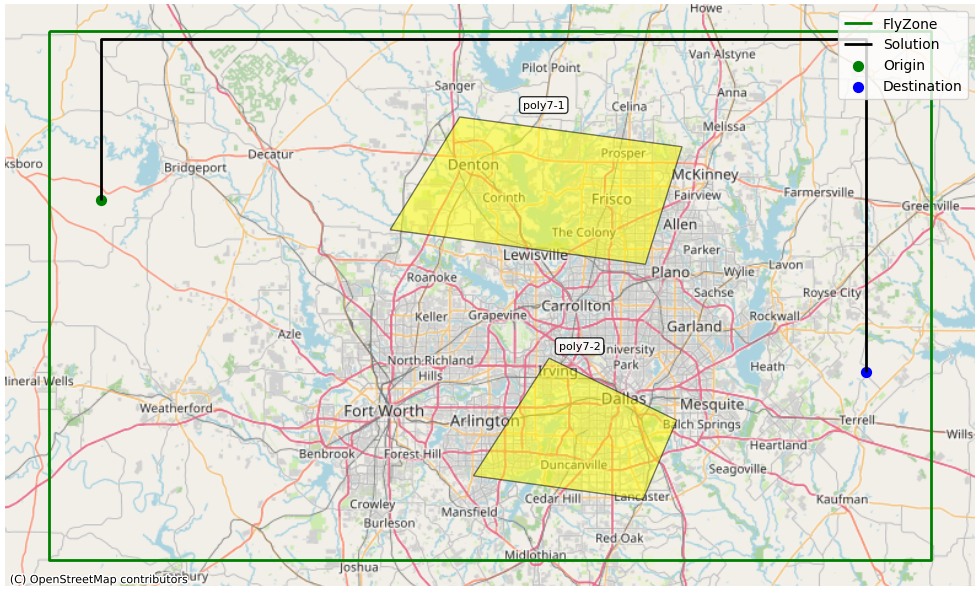}
  \caption{Easy scenario, clearance preference. The generated image of the flight plan.}
  \label{fig:qual_easy}
\end{figure}

\subsection{Medium scenario, minimize distance preference}
\begin{inputbox}
{\color{syscolor}\textbf{SYS:}} You are a flight planner for an eVTOL aircraft. The user will provide you with hazardous polygon information and request a flight plan from an origin to a destination. You must generate a flight plan as a list of waypoints starting from the origin coordinate and ending at the destination coordinate while avoiding the hazardous polygons. Always include both the origin and destination points in your response. You can generate as many waypoints as necessary to avoid the polygons---more waypoints may lead to a smoother flight plan. You cannot fly outside of the flyzone. The best approach to find the optimal solution is as follows: (1) Identify the origin and destination points. (2) Identify the hazardous polygons and the flyzone. (3) IMPORTANT STEP: Generate waypoints that connect the origin to the destination while avoiding hazardous polygons and staying within the flyzone (they should not be on the flyzone's border either). You may generate more waypoints near the hazardous polygons to ensure that the line segments do not intersect with the hazardous polygons. Ensure that the flight plan connecting the waypoints are aligned with the human preference. YOU NEED TO INCLUDE AT LEAST 4 DECIMAL POINTS FOR WAYPOINT COORDINATES. (4) The line segments connecting the waypoints should not have sharp angles (recommended). (5) Ensure that the line segments do not intersect with the hazardous polygons. (6) If any of the line segments intersect with the polygons, modify the corresponding waypoints so the new line segment does not intersect the polygon.

\smallskip
{\color{usercolor}\textbf{USER:}} Now you have to generate a flight plan avoiding the hazardous polygons for the following problem. Coordinates for `FlyZone': (\mbox{$-$98.082},\,33.465), (\mbox{$-$98.082},\,32.473), (\mbox{$-$96.110},\,32.473), (\mbox{$-$96.110},\,33.465). Coordinates for `poly4-1': (\mbox{$-$97.497},\,33.306), (\mbox{$-$97.254},\,33.140), (\mbox{$-$97.039},\,33.329), (\mbox{$-$97.181},\,33.436). Coordinates for `poly4-2': (\mbox{$-$97.094},\,32.993), (\mbox{$-$96.871},\,32.867), (\mbox{$-$96.601},\,33.051), (\mbox{$-$96.749},\,33.185). Coordinates for `poly4-3': (\mbox{$-$97.514},\,32.640), (\mbox{$-$97.263},\,32.632), (\mbox{$-$97.433},\,33.063), (\mbox{$-$97.720},\,32.985). Coordinates for `poly4-4': (\mbox{$-$97.022},\,32.547), (\mbox{$-$96.739},\,32.596), (\mbox{$-$96.769},\,32.706), (\mbox{$-$97.021},\,32.635). Coordinates for `Origin3': (\mbox{$-$97.966},\,32.893). Coordinates for `Destination4': (\mbox{$-$96.255},\,32.826). Human preference: Minimize total flight distance.

\smallskip
{\color{operatorcolor}\textbf{OPERATOR:}} ``Minimize total flight distance.''
\end{inputbox}

\begin{frameresponse}
\textbf{Waypoints:} (32.8931,\,\mbox{$-$97.9655}) $\to$ (32.6300,\,\mbox{$-$97.5137}) $\to$ (32.8261,\,\mbox{$-$96.2548}).

\smallskip
\textbf{Reasoning:} ``The direct path from origin to destination intersects poly4-3. To avoid it, we add a waypoint just south of poly4-3's bottom-left vertex, ensuring the flight path stays clear of all hazardous polygons and within the flyzone.''

\end{frameresponse}

\begin{coachreview}
\textbf{Valid:} False \quad \textbf{Waypoints outside flyzone:} \{\} \quad \textbf{Violated polygons:} [\texttt{poly4-3}]

\smallskip
\textbf{Coach Verdict:} {[NOT ALIGNED]}

\smallskip
\textbf{Evaluation:} Path is invalid---alignment not evaluated.
\end{coachreview}

\begin{figure}[!h]
  \centering
  \includegraphics[width=0.6\linewidth]{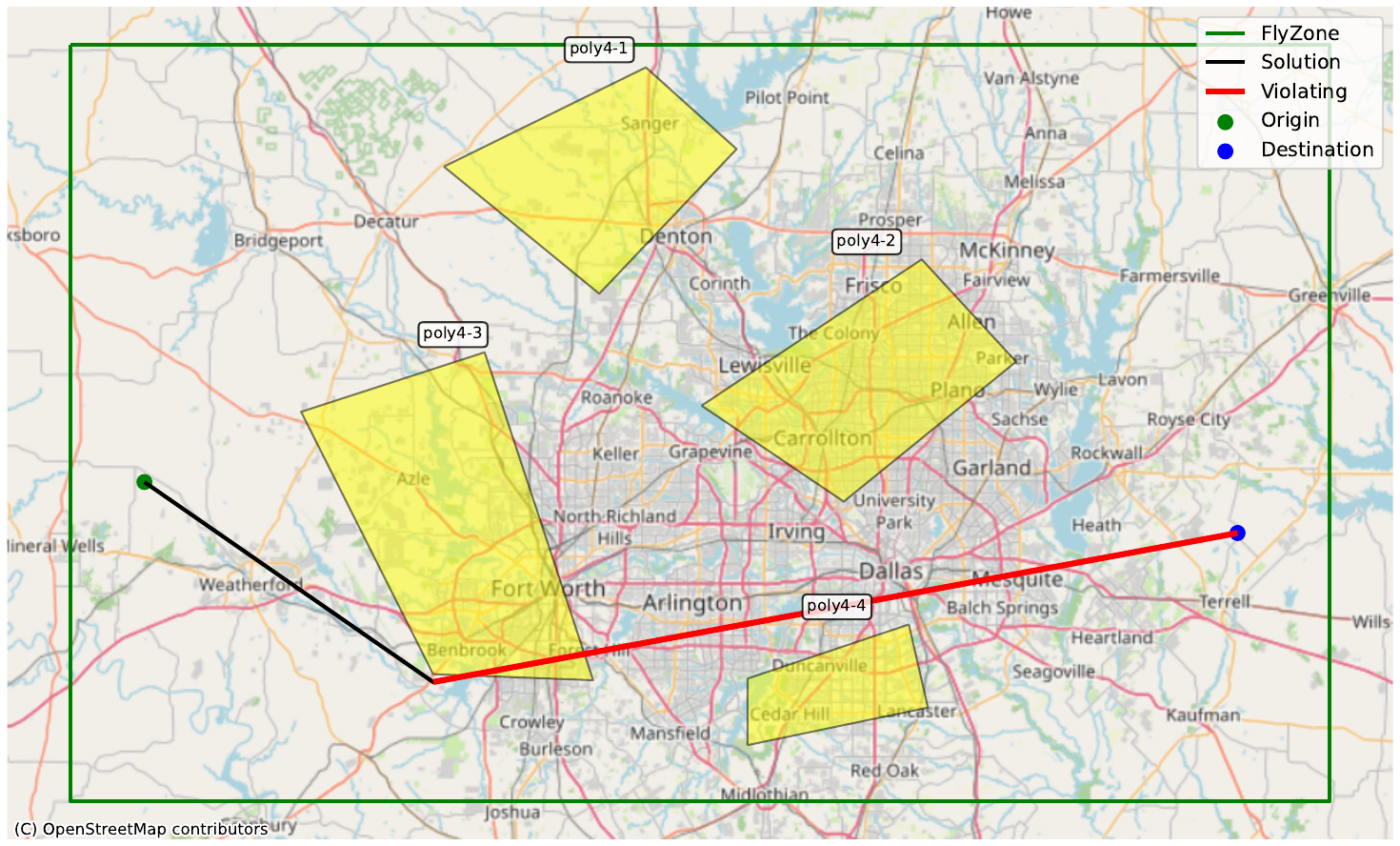}
  \caption{Medium scenario, minimize distance preference. The generated image of the flight plan.}
  \label{fig:qual_medium_distance}
\end{figure}

\subsection{Hard scenario, minimize waypoints preference}
\begin{inputbox}
{\color{syscolor}\textbf{SYS:}} You are a flight planner for an eVTOL aircraft. The user will provide you with hazardous polygon information and request a flight plan from an origin to a destination. You must generate a flight plan as a list of waypoints starting from the origin coordinate and ending at the destination coordinate while avoiding the hazardous polygons. Always include both the origin and destination points in your response. You can generate as many waypoints as necessary to avoid the polygons---more waypoints may lead to a smoother flight plan. You cannot fly outside of the flyzone. The best approach to find the optimal solution is as follows: (1) Identify the origin and destination points. (2) Identify the hazardous polygons and the flyzone. (3) IMPORTANT STEP: Generate waypoints that connect the origin to the destination while avoiding hazardous polygons and staying within the flyzone (they should not be on the flyzone's border either). You may generate more waypoints near the hazardous polygons to ensure that the line segments do not intersect with the hazardous polygons. Ensure that the flight plan connecting the waypoints are aligned with the human preference. YOU NEED TO INCLUDE AT LEAST 4 DECIMAL POINTS FOR WAYPOINT COORDINATES. (4) The line segments connecting the waypoints should not have sharp angles (recommended). (5) Ensure that the line segments do not intersect with the hazardous polygons. (6) If any of the line segments intersect with the polygons, modify the corresponding waypoints so the new line segment does not intersect the polygon.

\smallskip
{\color{usercolor}\textbf{USER:}} Now you have to generate a flight plan avoiding the hazardous polygons for the following problem. Coordinates for `FlyZone': (\mbox{$-$98.082},\,33.465), (\mbox{$-$98.082},\,32.473), (\mbox{$-$96.110},\,32.473), (\mbox{$-$96.110},\,33.465). Coordinates for `poly1-1': (\mbox{$-$97.584},\,33.194), (\mbox{$-$97.241},\,33.081), (\mbox{$-$97.104},\,33.375), (\mbox{$-$97.263},\,33.425). Coordinates for `poly1-2': (\mbox{$-$97.095},\,33.035), (\mbox{$-$96.773},\,32.814), (\mbox{$-$96.434},\,33.021), (\mbox{$-$96.586},\,33.346). Coordinates for `poly1-3': (\mbox{$-$97.643},\,32.826), (\mbox{$-$97.510},\,32.718), (\mbox{$-$97.248},\,32.905), (\mbox{$-$97.416},\,33.049). Coordinates for `poly1-4': (\mbox{$-$97.309},\,32.612), (\mbox{$-$97.208},\,32.625), (\mbox{$-$97.173},\,32.704), (\mbox{$-$97.265},\,32.694). Coordinates for `poly1-5': (\mbox{$-$97.030},\,32.644), (\mbox{$-$96.897},\,32.625), (\mbox{$-$96.898},\,32.747), (\mbox{$-$97.051},\,32.748). Coordinates for `poly1-6': (\mbox{$-$96.559},\,32.635), (\mbox{$-$96.507},\,32.686), (\mbox{$-$96.571},\,32.768), (\mbox{$-$96.629},\,32.691). Coordinates for `poly1-7': (\mbox{$-$96.765},\,33.428), (\mbox{$-$96.954},\,33.415), (\mbox{$-$96.947},\,33.262), (\mbox{$-$96.779},\,33.324). Coordinates for `Origin1': (\mbox{$-$97.970},\,33.359). Coordinates for `Destination1': (\mbox{$-$96.256},\,33.415). Human preference: Minimize the number of waypoints---keep the path as simple as possible. Previous solutions with coach reviews for similar problems. Example~1: Human preference: Propose the best flight plan from origin to destination avoiding hazardous polygons. Solution waypoints: (\mbox{$-$97.970},\,33.359) $\to$ (\mbox{$-$97.200},\,33.460) $\to$ (\mbox{$-$96.256},\,33.415). Valid: True. Waypoints outside flyzone: \{\}. Violated polygons: [].

\smallskip
{\color{operatorcolor}\textbf{OPERATOR:}} ``Minimize the number of waypoints---keep the path as simple as possible.''
\end{inputbox}

\begin{frameresponse}
\textbf{Waypoints:} (33.3594,\,\mbox{$-$97.9705}) $\to$ (33.4654,\,\mbox{$-$97.2000}) $\to$ (33.4155,\,\mbox{$-$96.2559}).

\smallskip
\textbf{Reasoning:} ``The flight plan starts at the origin, proceeds to an intermediate waypoint set at a high latitude ($33.4654$°N) to safely avoid all hazardous polygons, particularly poly1-1 and poly1-7, and then continues to the destination. This path minimizes the number of waypoints as per human preference while ensuring the aircraft stays within the flyzone and avoids polygon intersections by flying above the obstacles.''
\end{frameresponse}

\begin{coachreview}
\textbf{Valid:} True \quad \textbf{Waypoints outside flyzone:} \{\} \quad \textbf{Violated polygons:} {[}{]}

\smallskip
\textbf{Coach Verdict:} {[ALIGNED]}

\smallskip
\textbf{Evaluation:} Aligned with the preference: the route uses only one intermediate waypoint, keeping the path to two long segments while avoiding hazards.
\end{coachreview}

\begin{figure}[!h]
  \centering
  \includegraphics[width=0.6\linewidth]{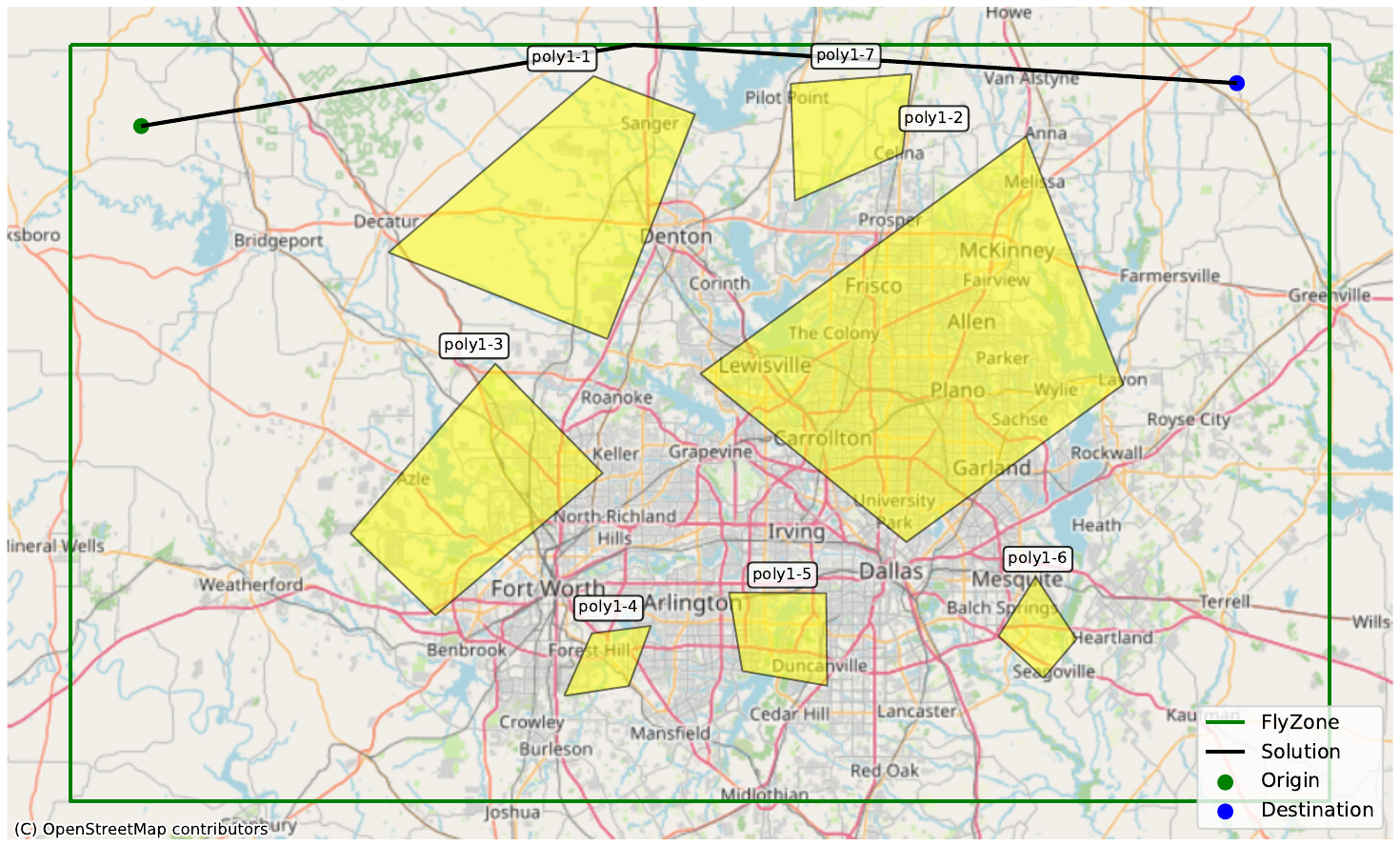}
  \caption{Hard scenario, minimize waypoints preference. The generated image of the flight plan.}
  \label{fig:qual_hard_wp}
\end{figure}

\section{Validity by Difficulty Level}\label{app:validity_difficulty}

\begin{figure}[!h]
  \centering
  \includegraphics[width=0.8\linewidth]{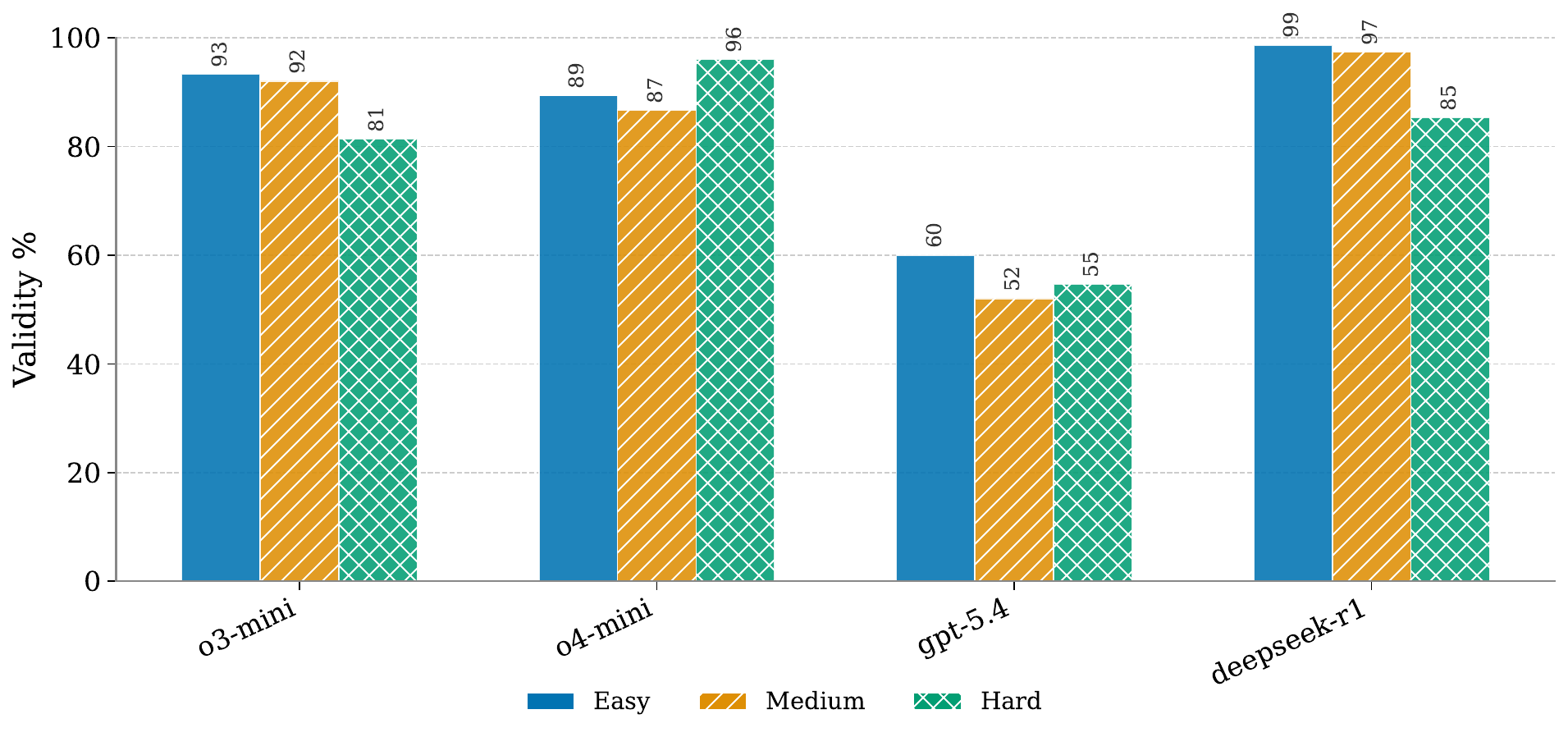}
  \caption{Plan validity (\%) of the full FRAMe system (+RAG+Coach) broken down by difficulty level (Easy, Medium, Hard) for each planner. The relationship between difficulty and validity is model-specific: DeepSeek-R1 peaks on Easy/Medium (98.7\%/97.3\%) while o4-mini peaks on Hard (96.0\%), and GPT-5.4 remains consistently the weakest planner across all difficulty levels (52--60\%).}
  \label{fig:validity_full}
\end{figure}

Figure~\ref{fig:validity_full} breaks down the full system (+RAG+Coach) by difficulty level and reveals that the relationship between geometric complexity and validity is model-specific rather than uniform. DeepSeek-R1 achieves the highest validity on Easy and Medium scenarios (98.7\% and 97.3\%, respectively) but drops to 85.3\% on Hard, suggesting that its strong internal reasoning becomes less reliable when navigating seven overlapping no-fly zones simultaneously. o4-mini exhibits the opposite pattern: its Hard validity (96.0\%) exceeds its Easy (89.3\%) and Medium (86.7\%) rates, indicating that richer geometric structure helps anchor this model's planning decisions rather than overwhelm them. o3-mini shows a monotone decline from Easy (93.3\%) through Medium (92.0\%) to Hard (81.3\%), consistent with a planner that handles moderate complexity well but degrades under the densest airspace configurations. GPT-5.4 remains consistently the weakest planner across all difficulty levels (52--60\%), with no clear sensitivity to difficulty; the coach partially compensates for its lower intrinsic planning capability but cannot fully bridge the gap to the other three models.

\section{Per-Preference Capture Matrix}\label{app:pref_matrix}
\begin{figure}[!h]
  \centering
  \includegraphics[width=0.9\linewidth]{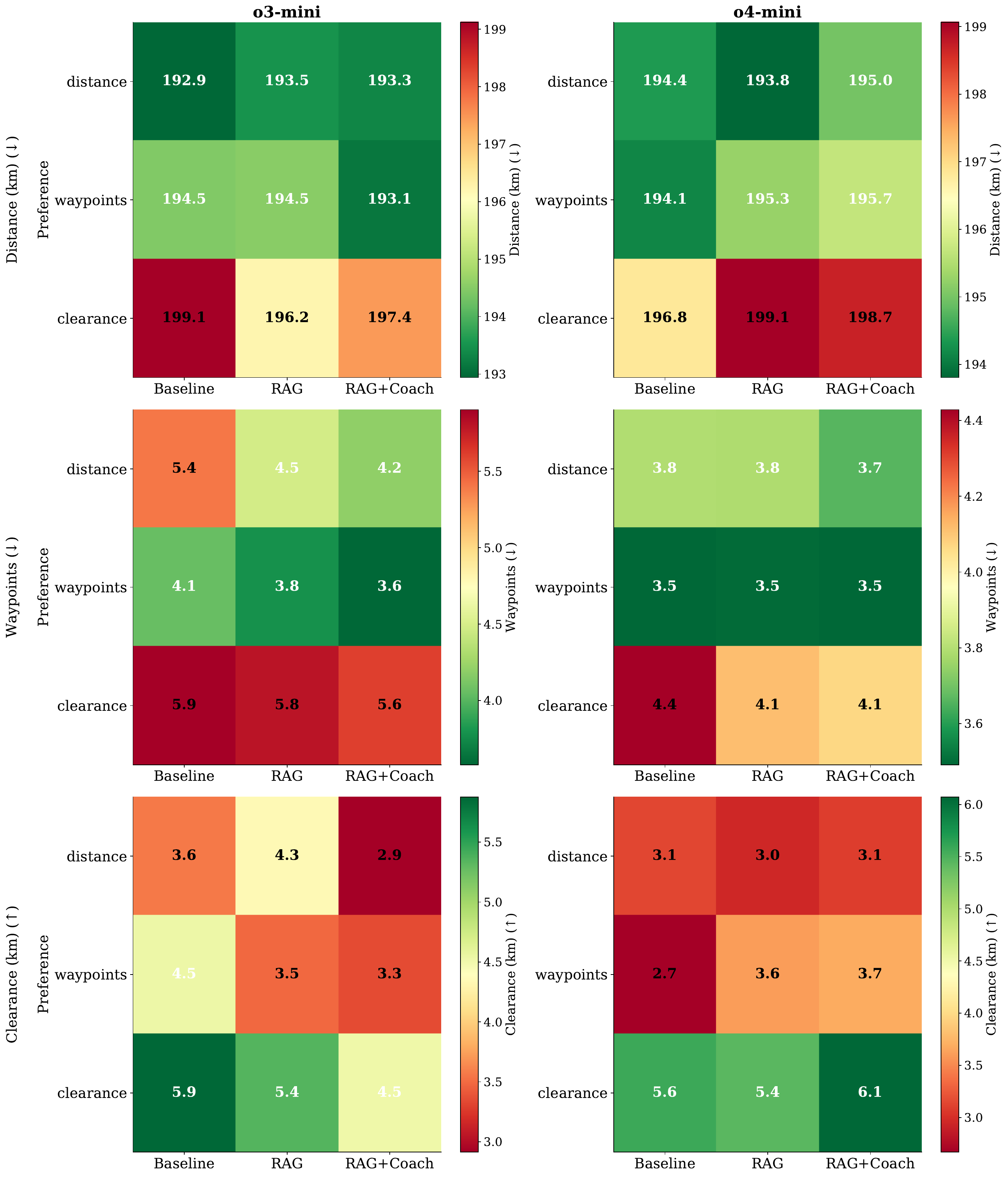}
  \caption{Mean metric values by preference and condition for each model. Each cell reports the average over valid plans (o3-mini and o4-mini).}
  \label{fig:metrics1}
\end{figure}

\begin{figure}[!h]
  \centering
  \includegraphics[width=0.9\linewidth]{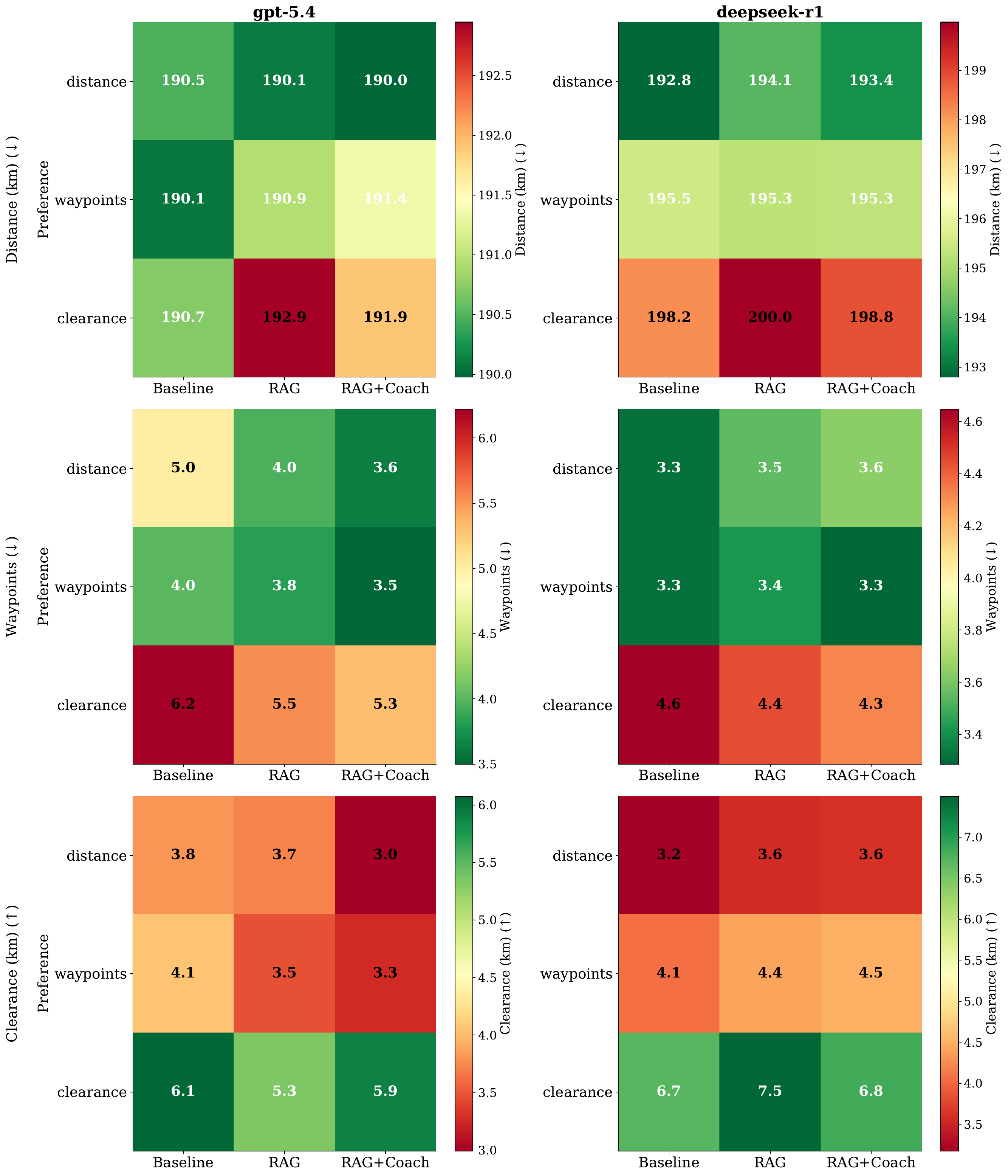}
  \caption{Mean metric values by preference and condition for each model. Each cell reports the average over valid plans(gpt-5.4 and deepseek-r1).}
  \label{fig:metrics2}
\end{figure}

Figures~\ref{fig:metrics1} and~\ref{fig:metrics2} show the full per-preference breakdown of all three metrics (distance, waypoints, clearance) across the three conditions (Baseline, +RAG, +RAG+Coach) for each model. Each cell reports the mean value over valid plans only. Reading across a row shows how a given metric changes as augmentation increases; reading down a column shows how different preferences affect the same metric under the same condition. The diagonal entries (where the preference matches the metric, e.g.\ distance metric under the distance preference) are the primary preference-capture signal: a diagonal cell that moves in the desired direction (lower distance or waypoints, higher clearance) relative to its Baseline indicates that the corresponding condition captures the active preference for that model. Off-diagonal entries are informative as side-effect probes---e.g.\ asking for clearance can lengthen distance, or asking for fewer waypoints can reduce clearance---and should be read as costs incurred to satisfy a different active preference rather than as failures of preference capture.

\end{document}